\documentclass{article}
\usepackage{spconf,amsmath,graphicx}
\usepackage{booktabs,multirow}
\usepackage{amssymb} 

\def\figref#1{Figure~\ref{fig:#1}}
\def\tabref#1{Table~\ref{tab:#1}}

\newcommand{\etal}{\textit{et al}. }

\title{FOCUSING ON TARGETS FOR IMPROVING WEAKLY SUPERVISED VISUAL GROUNDING}
\name{Viet-Quoc Pham \qquad Nao Mishima}
\address{Corporate Research and Development Center, Toshiba Corporation}
\begin{document}
\ninept
\maketitle
\begin{abstract}
Weakly supervised visual grounding aims to predict the region in an image that corresponds to a specific linguistic query, where the mapping between the target object and query is unknown in the training stage. The state-of-the-art method uses a vision language pre-training model to acquire heatmaps from Grad-CAM, which matches every query word with an image region, and uses the combined heatmap to rank the region proposals. In this paper, we propose two simple but efficient methods for improving this approach. First, we propose a target-aware cropping approach to encourage the model to learn both object and scene level semantic representations. Second, we apply dependency parsing to extract words related to the target object, and then put emphasis on these words in the heatmap combination. Our method surpasses the previous SOTA methods on RefCOCO, RefCOCO+, and RefCOCOg by a notable margin.
\end{abstract}

\section{Introduction}
Visual grounding (VG) (also known as phrase localization, referring expression comprehension, and natural language object retrieval) is the task of finding the corresponding location of a referential object given a textual query and an image. VG fills an important role at the intersection of computer vision and natural language processing. 

\begin{figure}[t]
 \centering
 \includegraphics[width=\linewidth]{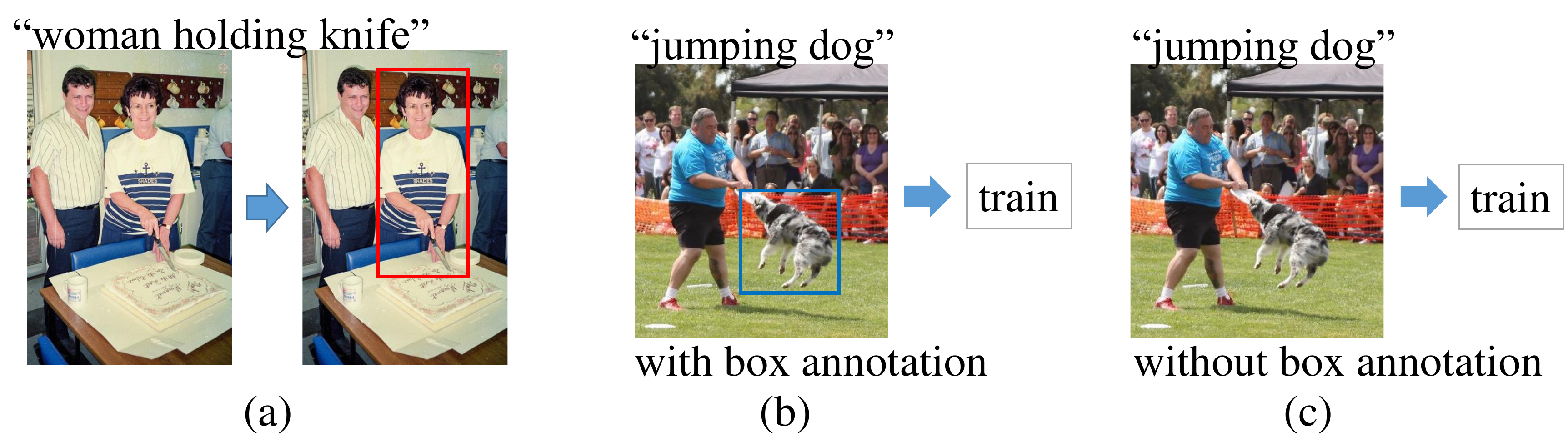}
 \vspace{-7mm}
 \caption{(a) Visual grounding trained in a (b) fully supervised manner and (c) weakly supervised manner.} 
 \vspace{-3mm}
 \label{fig:vg}
\end{figure}

Existing methods for VG can be divided into two categories: fully supervised VG and weakly supervised VG (see \figref{vg}). Fully supervised VG methods require annotation of target object rectangles in the training stage. These rectangle annotations explicitly draw the connection between the input query and its corresponding object region in the image. However, this kind of manual annotation is expensive, particularly if we want to increase the versatility of the model to unknown categories.

In this paper, we address the task of weakly supervised VG, where only the image--query pairs are used for training. In this setting, the mapping between the query and the target object cannot be learned directly from the training data.

Weakly supervised VG is a new research topic with few related studies and not yet fully explored. Early methods for weakly supervised VG are based on multiple-instance learning or text reconstruction. Methods that are based on multiple-instance learning \cite{Gao2019,Mithun2019,Datta2019} collect positive samples from input image--text pairs and negative samples from unmatched pairs, and directly learn the image--text alignment. Reconstruction-based methods \cite{Song2020,Lin2020,Liu2019,Rohrbach2016} aim to reconstruct the text query from visual features during training and use the intermediate results, such as attention weights, to localize the target proposal during inference. However, these methods have several shortcomings, particularly their low accuracy in comparison with that of supervised VG approaches.

Recent approaches \cite{Li2021,Zeng2022} use vision and language pre-training (VLP), which learns multi-modal representations from large-scale image-text pairs, to improve downstream vision-and-language tasks, including weakly supervised VG. They introduce a contrastive loss to align the image and text representations through cross-modal attention, which enables more grounded vision and language representation learning. During inference, they use the VLP model to acquire heatmaps from Grad-CAM \cite{Selvaraju2017}, which matches every query word with an image region, and uses the combined heatmap to rank the region proposals. The state-of-the-art method X-VLM \cite{Zeng2022} has nearly doubled the accuracy of earlier methods for weakly supervised VG, narrowing the gap in performance between weakly and fully supervised VG. This paper focuses on these VLP-based approaches and proposes two methods to improve their performance in weakly supervised VG.

First, we propose a target-aware cropping approach to encourage the model to learn both object- and scene-level semantic representations. Cropping is a powerful and commonly used method of data augmentation, and is useful when the object of interest occupies most of the image and is largely centered within the image. This assumption holds for some datasets such as ImageNet \cite{Deng2009}. However, in the case of datasets used in weakly supervised VG, the objects of interest are small relative to the image size and are rarely centered. Our experiments show that the default random cropping method leads to a reduction in performance of weakly supervised VG. Inspired by the idea of object-aware cropping for self-supervised learning in \cite{Mishra2021}, our cropping approach focuses on the referential targets by replacing the random crops by crops obtained from heatmaps acquired from Grad-CAM.

Second, we introduce dependency parsing into the inference stage. Previous methods \cite{Li2021,Zeng2022} deal with all query words in the same way and make no distinction between main (or target) objects and sub-objects, often resulting in false detection. Dependency parsing is the process of analyzing the grammatical structure in a sentence and finding related words as well as the type of relationship between them \cite{Kubler2009}. In our method, we use dependency parsing to extract words related to the target object, and emphasize these words in the heatmap combination.

We summarize our contributions in the following points:
\begin{itemize}
\item We propose a new target-aware method of cropping during the training stage, and introduce dependency parsing into the inference stage. The proposed two methods improve the baseline methods \cite{Li2021,Zeng2022} by putting emphasis on the visual and textual contents of the referential object.
\item Experiments on three benchmark datasets RefCOCO \cite{RefCOCOplus}, RefCOCO+ \cite{RefCOCOplus} and RefCOCOg \cite{Mao2016,Nagaraja2016} show that the combination of the two proposed methods achieves state-of-the-art results in the weakly supervised VG task. For RefCOCOg, we surpass the previous method by 5 points.
\item Our methods are easy to implement, and can reuse the existing vision language pre-trained models \cite{Li2021,Zeng2022}, which require high-end GPUs for training.
\end{itemize}

\section{Related Work}
\noindent {\bf Visual Grounding:} Existing methods for VG can be divided into two categories: fully supervised VG and weakly supervised VG. Fully supervised VG approaches often train a model to directly map the input query to the target region \cite{Yu2018,Yang2019,Yang2020,Deng2021,Kamath2021}. CMRIN \cite{Yang2019} converts the pair of input query and image
into a language-guided visual relation graph which captures multi-modal semantic contexts with multi-order relationships. MDETR \cite{Kamath2021} pre-trains the transformer-based network on 1.3 million text-image pairs, mined from pre-existing multi-modal datasets that have explicit alignment between phrases in text and image objects. Early methods for weakly supervised VG use multiple-instance learning or text reconstruction to train the mapping model \cite{Gao2019,Mithun2019,Datta2019,Song2020,Lin2020,Liu2019,Rohrbach2016}. ARN \cite{Liu2019} models the mapping between an image proposal and query upon the subject, location, and context information through adaptive grounding and collaborative reconstruction. CCL \cite{Zhang2020} proposes a counterfactual contrastive learning, which develops sufficient contrastive training between positive and negative results. Recently, Wang \etal \cite{Wang2021} use a generic object detector at training time, and propose a contrastive learning framework that accounts for both region--phrase and image--sentence matching. A main limitation of their work is the need of a generic object detector to cover most of the object classes during training. Their problem setting is quite different from ours. Besides, they did not report results on standard VG benchmark datasets RefCOCO , RefCOCO+, and RefCOCOg.

\noindent {\bf Vision Language Pre-training (VLP):} Recently, the joint pre-training of models for vision and language tasks has attracted intense interest \cite{Tan2019,Chen2020,Li2021,Zeng2022}. A common approach is to view regions and words as tokens for their respective domains and to pre-train a variant of BERT \cite{Devlin2019} for masked token prediction. The models are pre-trained with large amounts of image--sentence pairs by using several diverse representative pre-training tasks, and then are finetuned to downstream tasks, such as visual question answering (VQA) and VG. CLIP \cite {Radford2021} is pretrained on 400M image-caption pairs collected from the web and achieves impressive zero-shot image classification performance on a variety of visual domains. Subramanian \etal \cite{Subramanian2022} propose a zero-shot VG method ReCLIP, which uses CLIP to score object proposals and handle spatial relations between objects.

\noindent {\bf Weakly Supervised VG with VLP:} ALBEF \cite{Li2021} introduces a contrastive loss to align the image and text representations before fusing them through cross-modal attention. ALBEF and its extended model X-VLM \cite{Zeng2022} improve previous VLP models, achieving state-of-the-art performance on multiple downstream tasks, including cross-modal retrieval and weakly supervised VG.


ALBEF contains an image encoder, a text encoder, and a cross-modal encoder. The image encoder uses the 12-layer visual transformer ViT-B/16 \cite{Kolesnikov2021}. The text encoder and the cross-modal encoder use a 6-layer transformer \cite{Vaswani2017}. X-VLM extends ALBEF by adding image sub-regions with associated texts to the input. It enriches the training data, and learns multi-grained alignments between vision and language in pre-training.

\begin{figure}[t]
 \centering
 \includegraphics[width=\linewidth]{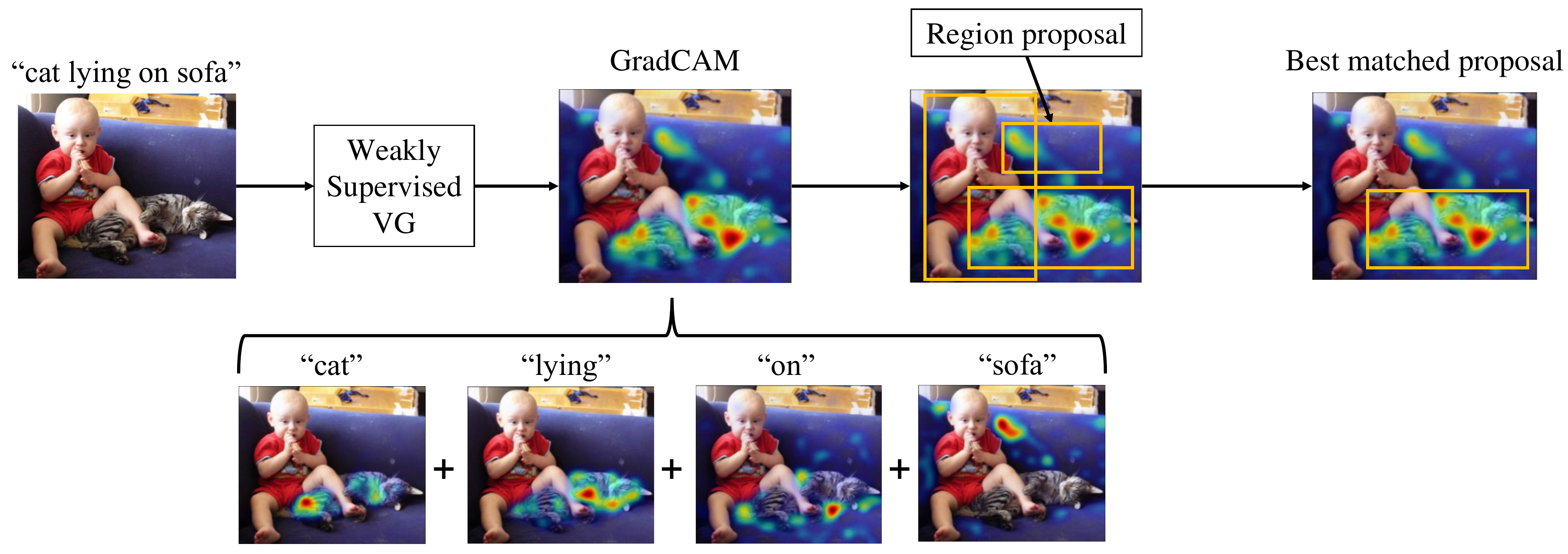}
 \vspace{-5mm}
 \caption{Inference process for weakly supervised VG.} 
 \vspace{-3mm}
 \label{fig:vg_infer}
\end{figure}

For the weakly supervised VG downstream task, the model is finetuned by using only image--text supervision to jointly optimize the image--text contrastive loss and the matching loss \cite{Li2021,Zeng2022}. The inference process for weakly supervised VG of ALBEF and X-VLM is shown in \figref{vg_infer}. For each input text token, a Grad-CAM \cite{Selvaraju2017} is computed on the cross-attention map in the third layer of the cross-modal encoder, where the gradient is acquired by maximizing the image-text matching score. The heatmap of the input query $H(q)$ is computed by averaging the Grad-CAMs across all input text tokens:
\begin{equation}
H(q)=\dfrac{1}{N}\sum_i{G(t_i)}
\label{eq:gradcam}
\vspace{-2mm}
\end{equation}
where $G(t_i)$ is the Grad-CAM of the token $t_i$, and $N$ is the number of input tokens. Finally, object proposals are detected by a region proposal network \cite{Ren2015}, and are ranked using the score which is the averaged heatmap value in each object proposal area. The output result is the proposal with the highest score.

\section{Proposed Methods}
In this section, we propose a new cropping method in the training stage, and introduce dependency parsing into the inference stage. The two methods improve the baseline methods \cite{Li2021,Zeng2022} by putting emphasis on the visual and textual contents of the referential objects. In this work, we use the two baselines as examples to illustrate our methods. Our improvement is not limited to these baselines and can be applied to any VLP models using image--text matching.

\subsection{Target-aware Cropping}
Cropping is a powerful and commonly used method of data augmentation. Cropping then resizing forces the representation to focus on different parts of an object with varying aspect ratios, making it robust to scaling and occlusion. The assumption behind the use of cropping is that the object of interest occupies most of the image and is quite centered within the image. This assumption holds for classification-oriented datasets such as ImageNet \cite{Deng2009}. However, in the case of weakly supervised VG datasets, the objects of interest are small relative to the image size and are rarely centered. Our experiments show that the default random cropping method causes a reduction in performance of weakly supervised VG. To solve this problem, we propose a target-aware cropping approach that focuses on the referential targets by replacing the random crops by crops obtained from the heatmaps acquired from Grad-CAM \cite{Selvaraju2017}.

\begin{figure}[t]
 \centering
 \includegraphics[width=\linewidth]{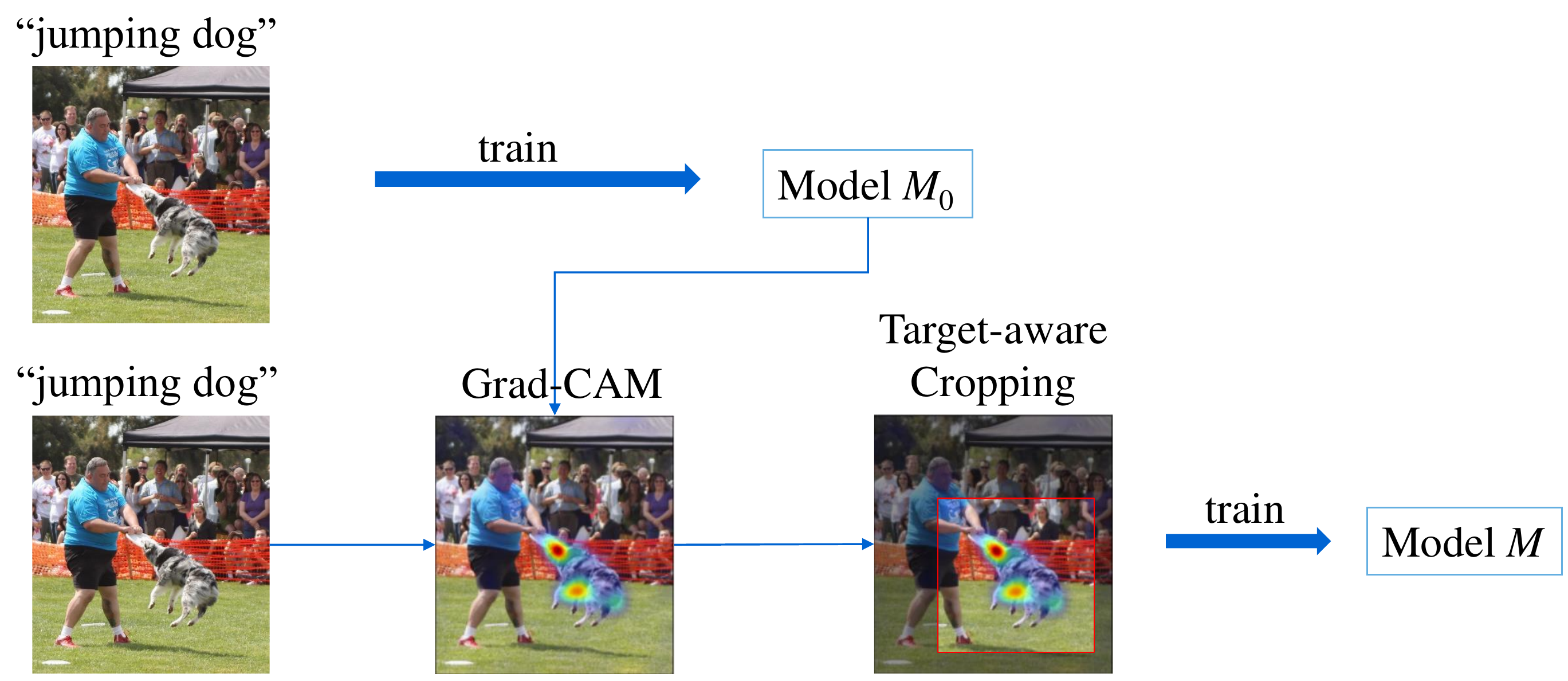}
 \vspace{-5mm}
 \caption{Training process using target-aware cropping.} 
 \label{fig:vg_crop}
\end{figure}

\begin{figure}[t]
 \centering
 \includegraphics[width=0.8\linewidth]{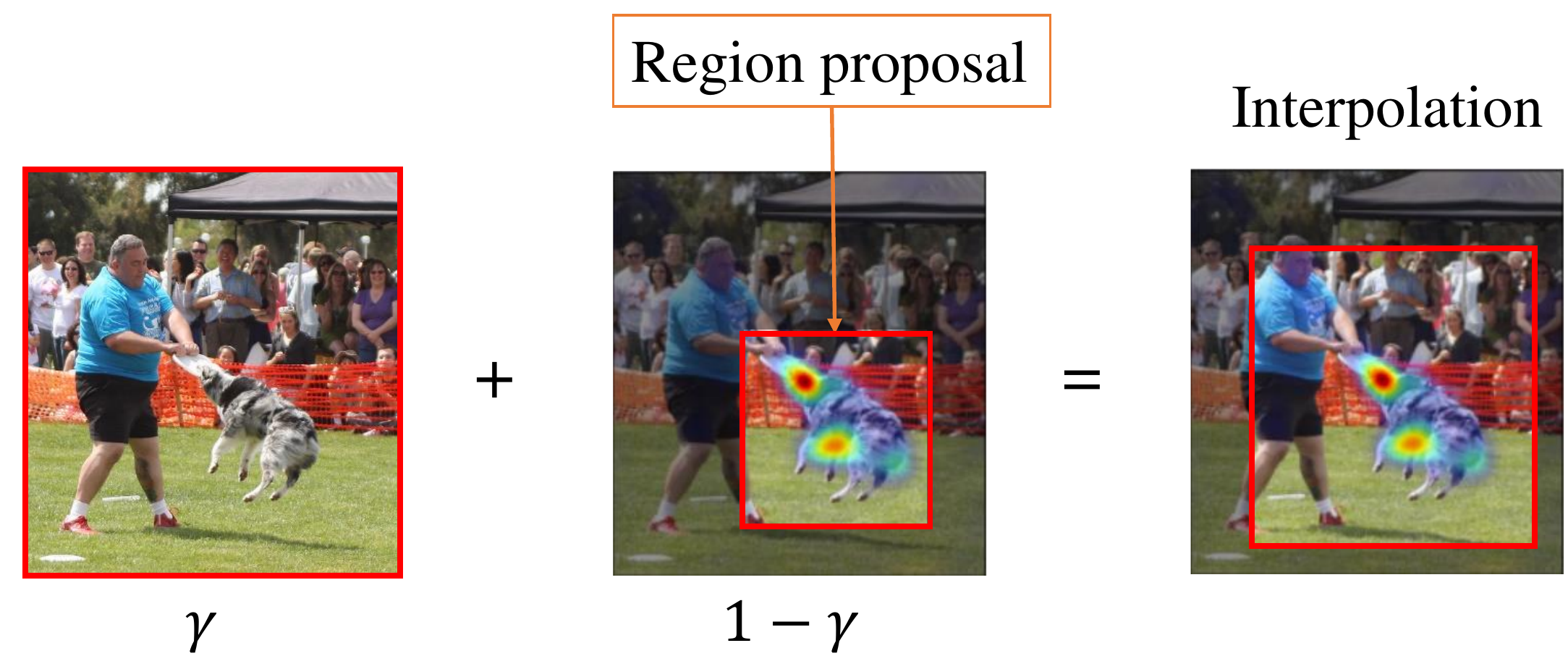}
 \vspace{-3mm}
 \caption{Illustration of the interpolation of the bounding boxes.} 
 \vspace{-3mm}
 \label{fig:interpolate}
\end{figure}

The training process using target-aware cropping is shown in \figref{vg_crop}. Given a starting model $M_0$, which is a pre-trained model or a finetuned model of \cite{Li2021,Zeng2022}, we aim to train a model $M$ using $M_0$ with target-aware cropping. For each training sample, we first use $M_0$ to roughly detect the referential object region of this sample. This process includes acquiring the heatmap from Grad-CAM and detecting the most matched proposal region (see \figref{vg_infer}). As this process may cause false positives and the detected region may be too tightly wrapped around the object, we apply here a simple interpolation to acquire the final cropped region, as follows:
\begin{equation}
r = \gamma r_0 + (1 - \gamma)r_1
\label{eq:interpolate}
\vspace{-1mm}
\end{equation}
where $r_0$, $r_1$ and $r$ are the bounding boxes\footnote{$r=(x_1,y_1,x_2,y_2)$ is a vector of coordinates.} of the whole image, the detected object region and the final cropped region, respectively. $\gamma$ is randomly generated in the range $[\gamma_{min},1]$, where $\gamma_{min}$ is the cropping parameter defining the cropping range. In VG, not only the object itself but also the interaction with the surroundings may be important, so interpolation with the whole image is effective. \figref{interpolate} shows an illustration of this interpolation. In the remaining step, we crop the image to $r$, then resize it to the original image size, before continuing training the model $M$.

\subsection{Dependency Parsing}
Previous methods \cite{Li2021,Zeng2022} deal with all query words in the same way (see Equation \ref{eq:gradcam}) and make no distinction between main objects (or referential targets) and sub-objects, often resulting in false detection. Observing the inference result of X-VLM \cite{Zeng2022} in \figref{quality_compare}(a) (``women under pink umbrella''), we can see that the sub-object ``umbrella'' dominates the main object ``women'' in the heatmap. The same problem can be observed in the heatmap of X-VLM in \figref{quality_compare}(c) (``horse that has woman with brown coat''). Such inappropriate heatmaps lead to false positives in the inference results.

\begin{figure}[t]
 \centering
 \includegraphics[width=0.8\linewidth]{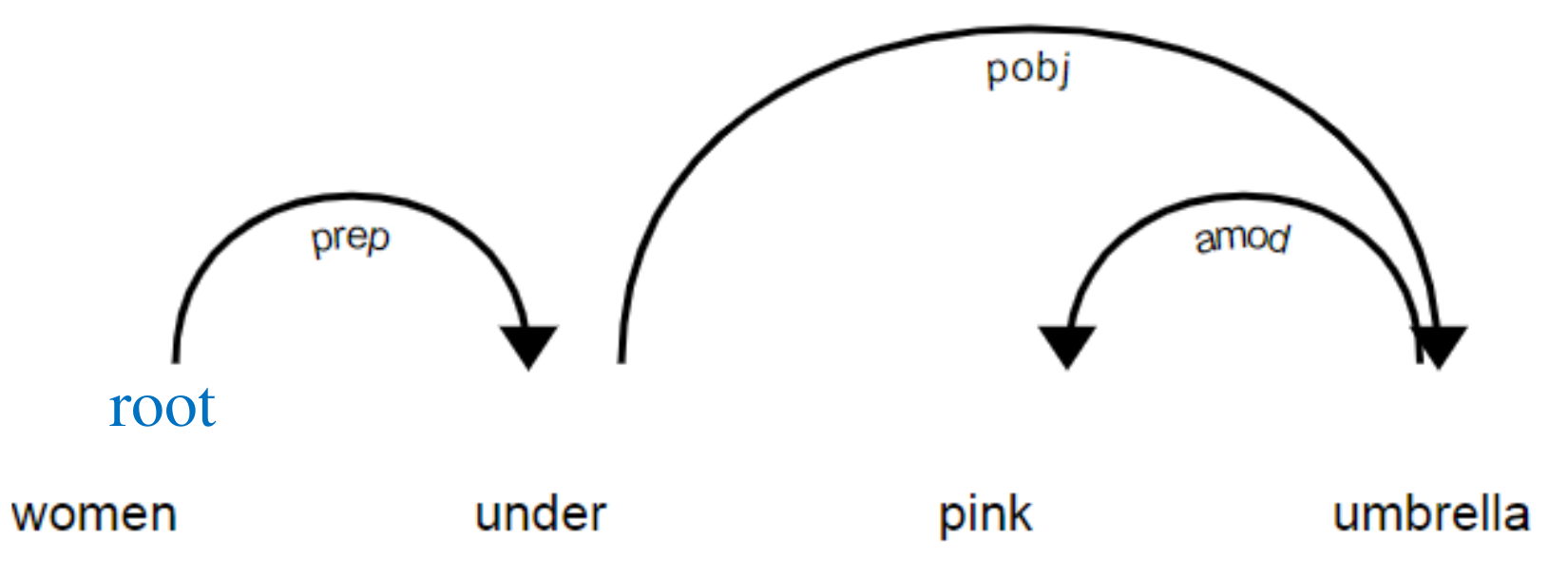}
 \vspace{-5mm}
 \caption{Example of dependency parsing using spaCy \cite{spacy} (prep: prepositional modifier, pobj: object of a preposition, amod: adjectival modifier). Each arc connects a parent node to a child node.}
 \vspace{-3mm}
 \label{fig:dp}
\end{figure}

To solve this problem, we introduce into the inference stage dependency parsing, which is the process of analyzing the grammatical structure in a sentence and finding related words as well as the type of relationship between them \cite{Kubler2009}. An example of dependency parsing using spaCy \cite{spacy} is shown in \figref{dp}. In this figure, a sentence is represented in a tree structure, where the relationships between each word in the sentence are expressed by directed arcs. A dependence tag attached to each arc indicates the relationship between the two words. 

Our proposed method re-defines the heatmap of the query $H(q)$ in Equation \ref{eq:gradcam} as a weighted average of the Grad-CAMs across all input text tokens
\begin{equation}
H(q)=\dfrac{1}{N}\sum_i{w_i G(t_i)}
\label{eq:our_gradcam}
\vspace{-2mm}
\end{equation}
where the weighting factor $w_i$ is computed using the dependency parsing results. Specifically, we first detect the root of the dependency parsing tree, which is the node without parents. The word at this root is expected to correspond to the main object of the query sentence. By denoting the input token of this word by $t_r$ indexed by $r$, we define the weighting factor $w_i$ as follows
\begin{equation}
w_i = 
	\begin{cases}
		1,			& \text{if }i \le r \\
		\alpha,	& \text{otherwise.}
	\end{cases}
\label{eq:weighting_factor}
\vspace{-2mm}
\end{equation}

We keep the default weight ($w_i=1$) for tokens appearing on the left of the root token $t_r$, and reduce the weights of the remaining tokens to $\alpha$ (where $0 \le \alpha < 1$). This is because words preceding the root are often nouns or adjectives modifying the main object, while words succeeding the root are often related to sub-objects. We put emphasis on words related to the main object to intensify its corresponding region in the heatmap.

\section{Experiments}
We conducted experiments on three standard benchmark datasets, RefCOCO \cite{RefCOCOplus}, RefCOCO+ \cite{RefCOCOplus} and RefCOCOg \cite{Mao2016,Nagaraja2016}. Each of RefCOCO, RefCOCO+, and RefCOCOg contains 19,994, 19,992, and 26,711 images from COCO dataset \cite{Lin2014}, with 50,000, 49,856, and 54,822 annotated objects and 142,209, 141,564, and 104,560 annotated expressions, respectively. Different from RefCOCO, the queries in RefCOCO+ are disallowed to use locations to describe the referents. We used {\it only image--text supervision} during training. The region proposal network \cite{Ren2015} was trained on COCO images excluding the val/test images in the three benchmarks, as in \cite{Yu2018}. For evaluation, we calculated the accuracy as the metric. If the intersection over union (IoU) between the output region and the ground truth is larger than 0.5, we regarded it as a correct grounding result.

\begin{table*}[t]
\centering
\caption{Comparison with previous methods on three benchmark datasets (U: UMD partition, G: Google partition). The last row shows the gains over the previous state-of-the-art method X-VLM \cite{Zeng2022}.} 
\label{tab:comparison}
\begin{tabular}{l|c|c|c|c|c|c|c|c|c|c}
\hline
\multirow{2}{*}{Methods} & \multirow{2}{*}{Venue} & \multicolumn{3}{c|}{RefCOCO} & \multicolumn{3}{c|}{RefCOCO+} & \multicolumn{3}{c}{RefCOCOg} \\
\cline{3-11}
 & & val & test A & test B & val & test A & test B & val (U) & test (U) & val (G) \\
\hline
ARN \cite{Liu2019} 		& ICCV'19 			& 32.17 & 35.35 & 30.28 & 32.78 & 34.35 & 32.13 & - & - & 33.09 \\
CCL \cite{Zhang2020} 	& NeurIPS'20 		& 34.78 & 37.64 & 32.59 & 34.29 & 36.91 & 33.56 & - & - & 34.92 \\
ReCLIP \cite{Subramanian2022} & ACL'22 & 54.04 & 58.60 & 49.54 & 55.07 & 60.47 & 47.41 & 60.85 & 61.05 & - \\
ALBEF \cite{Li2021} 		& NeurIPS'21 		& 55.92 & 64.57 & 45.95 & 58.46 & 65.89 & 46.25 & 58.34 & 56.78 & 60.22 \\
ALBEF+ours 					& 							& 56.82 & 66.63 & 46.32 & 59.94 & 68.97 & 47.68 & 62.28 & 62.53 & 65.42 \\
X-VLM \cite{Zeng2022} 	& ICML'22 			& 67.78 & 74.95 & 57.86 & 68.46 & 76.53 & 57.09 & 69.89 & 68.18 & 71.00 \\
X-VLM+ours 					& 							& \textbf{68.75} & \textbf{76.58} & \textbf{58.67} & \textbf{70.73} & \textbf{79.24} & \textbf{58.58} & \textbf{73.68} & \textbf{72.48} & \textbf{76.06} \\
\hline
gain									&							& 0.97 & 1.63 & 0.81 & 2.27 & 2.71 & 1.49 & 3.79 & 4.30 & \textbf{5.06} \\
\hline
\end{tabular}
\vspace{-5mm}
\end{table*}

We used ALBEF \cite{Li2021} and X-VLM \cite{Zeng2022} as our baselines. When training with target-aware cropping, we used the finetuned model provided by the authors\footnote{For X-VLM, we used the provided model finetuned from the 4M-based checkpoint (which was pre-trained from 4 million images).} as the starting model $M_0$. We followed the same strategy as ALBEF and trained the model for 5 epochs. For dependency parsing, we used spaCy \cite{spacy} in all experiments.

\tabref{comparison} shows the results on three benchmarks, where our method outperformed existing methods \cite{Liu2019,Zhang2020,Li2021,Zeng2022}. By introducing target-aware cropping and dependency parsing, we improved the accuracies of both the baselines ALBEF and X-VLM by a notable margin. We achieved the highest gain for RefCOCOg with 5.06 points. Compared to RefCOCO and RefCOCO+, the queries in RefCOCOg are much longer (8.4 words on average). This result shows that our method works especially well for complex queries.

For ablation study, we conducted experiments to verify the effectiveness of each component in our proposed methods. In this study, we report only experimental results with the baseline X-VLM on RefCOCO+. First, we show the efficiency of each proposed method in \tabref{efficiency}. Compared with the baseline, we saw consistent performance gains after adding each method. The combination of the two methods gave the best results.

\begin{table}[tb]
\centering
\caption{Efficiency of target-aware cropping (TC) and dependency parsing (DP).} 
\label{tab:efficiency}
\begin{tabular}{l|c|c|c|c|c}
\toprule
Methods & TC & DP & val & testA & testB \\
\midrule
X-VLM& & & 68.46 & 76.53 & 57.09 \\
X-VLM+TC & \checkmark & & 69.39 & 77.68 & 58.31 \\
X-VLM+DP & & \checkmark & 70.31 & 79.06 & 58.01 \\
X-VLM+all & \checkmark & \checkmark & \textbf{70.73} & \textbf{79.24} & \textbf{58.58} \\
\bottomrule
\end{tabular}
\vspace{-5mm}
\end{table}

To further verify the efficiency of target-aware cropping, we compared our cropping method with default cropping (with random sizes and aspect ratios) and the approach without cropping. \tabref{no_crop} shows that the default cropping resulted in a reduction in performance, while our cropping achieved notable gains.

\begin{table}[tb]
\centering
\caption{Comparison with default cropping and the approach without cropping.} 
\label{tab:no_crop}
\begin{tabular}{l|c|c|c}
\toprule
 & Default cropping & Without cropping & Our cropping \\
\midrule
val & 67.85 & 68.46 & \textbf{69.39} \\
testA & 75.69 & 76.53 & \textbf{77.68} \\
testB & 56.43 & 57.09 & \textbf{58.31} \\
\bottomrule
\end{tabular}
\vspace{-5mm}
\end{table}

\tabref{crop_gamma} studies the effect of the cropping parameter $\gamma_{min}$, which defines the cropping range. When $\gamma_{min}=1$, we performed no cropping at all, so that the same performance was achieved as the baseline. The best performance was achieved when $\gamma_{min}=0.5$.

\begin{table}[tb]
\centering
\caption{Ablative experiments of the cropping parameter $\gamma_{min}$.} 
\label{tab:crop_gamma}
\begin{tabular}{l|ccccc}
\toprule
$\gamma_{min}$ & 1.0 (baseline) & 0.7 & 0.5 & 0.2 & 0.0 \\
\midrule
val & 		68.46 & 68.92 & \textbf{69.39} & 69.05 & 68.74 \\
testA & 	76.53 & 76.81 & \textbf{77.68} & 77.12 & 77.05 \\
testB & 	57.09 & 58.09 & 58.31 & \textbf{58.91} & 58.52 \\
\bottomrule
\end{tabular}
\vspace{-3mm}
\end{table}

\tabref{weight_alpha} shows the effect of the weighting parameter for dependency parsing (see Equation \ref{eq:weighting_factor}). When $\alpha=1$, we did not modify the heatmap weighting factor, resulting in low accuracies. We observed that the best performance was achieved when $\alpha$ is near $0.1$, therefore we conducted experiments on $\sqrt{\alpha}$ to find its optimal value (here, $\sqrt{\alpha} = 0.4$ or $\alpha = 0.16$). 

\begin{table}[tb]
\centering
\caption{Ablative experiments of the weighting parameter $\alpha$. All models were trained with target-aware cropping $\gamma_{min}=0.5$.} 
\label{tab:weight_alpha}
\begin{tabular}{l|ccccc}
\toprule
$\sqrt{\alpha}$ & 1.0 & 0.8 & 0.6 & 0.4 & 0.2 \\
\midrule
val & 		69.39 & 69.90 & 70.44 & \textbf{70.73} & 70.57 \\
testA & 	77.68 & 78.34 & 78.89 & \textbf{79.24} & 79.10 \\
testB & 	58.31 & 58.34 & 58.40 & \textbf{58.58} & 58.23 \\
\bottomrule
\end{tabular}
\end{table}

\begin{figure}[t]
 \centering
 \includegraphics[width=\linewidth]{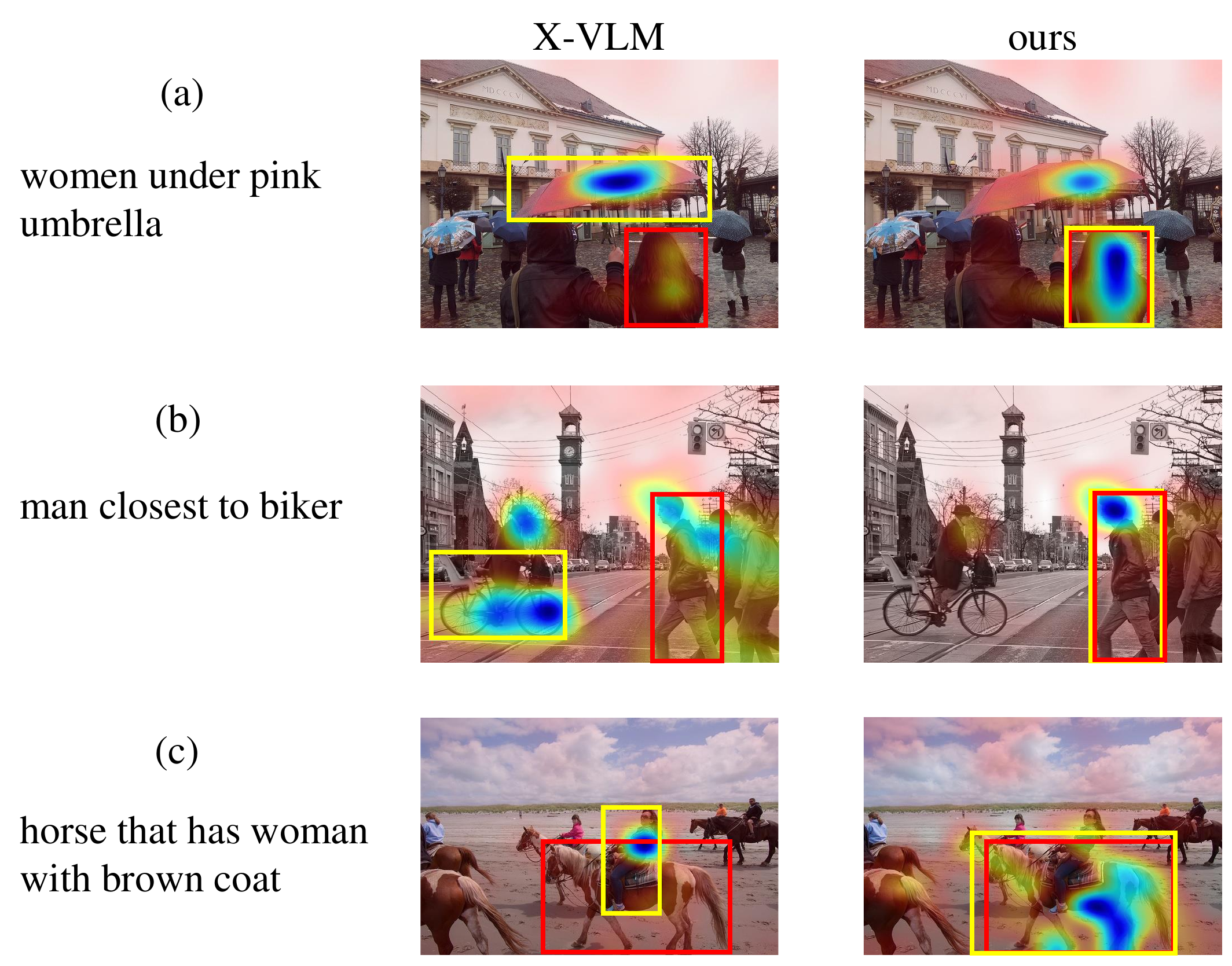}
 \vspace{-5mm}
 \caption{Inference results of X-VLM (middle column) and ours (right column). Computed heatmaps are displayed with detection results (in yellow boxes) and ground truth (in red boxes).} 
 \vspace{-2mm}
 \label{fig:quality_compare}
\end{figure}

We show the qualitative results of some examples from the RefCOCO+ val/test set, in comparison with X-VLM, in \figref{quality_compare}. We obtained better heatmaps that correctly capture main/sub object relationships. For instance, in \figref{quality_compare}(a), our modified heatmap could explain the scene correctly, where the main object ``women'' dominated the sub-object ``umbrella'' in the heatmap.

\section{Conclusion}
This paper addresses the problem of weakly supervised VG which yields great value for various applications with its low annotation costs. We propose two target-aware methods in both the training and inference stages for improving previous methods. Experimental results demonstrate that our methods provide a significant improvement in performance on three benchmarks. 

\vfill\pagebreak

\bibliographystyle{IEEEbib-abbrev}
\bibliography{ref}

\end{document}